\documentclass[11pt]{article}

\usepackage{geometry}
\usepackage{setspace}
\usepackage{caption} 
\usepackage{graphicx}

\usepackage{mathptmx}
\usepackage{amssymb}
\usepackage[T1]{fontenc}

\usepackage{url}
\usepackage[natbibapa]{apacite}

\usepackage{booktabs} 
\usepackage{tabularx} 
\usepackage{rotating} 
\usepackage[hidelinks]{hyperref} 
\usepackage[colorinlistoftodos]{todonotes}
\usepackage{cleveref} 
\usepackage{float}
\usepackage{xspace}

\newenvironment{biseabstract}{%
\begin{quote} \bf}
{\end{quote}}

\newenvironment{bisekeywords}{%
\begin{quote} \it \textbf{Keywords:}}
{\end{quote}}

\title{Generative AI}

\author{
Stefan Feuerriegel$^{1,*}$ and Jochen Hartmann$^{2}$ and Christian Janiesch$^{3}$ and Patrick Zschech$^{4}$\\
\\
\normalsize{$^{1}$LMU Munich \& Munich Center for Machine Learning, Geschwister-Scholl-Platz 1, 80539 Munich,}\\ \normalsize{feuerriegel@lmu.de} \\
\normalsize{$^{2}$Technical University of Munich, TUM School of Management, Arcisstr. 21, 80333 Munich,}\\ \normalsize{jochen.hartmann@tum.de}\\
\normalsize{$^{3}$TU Dortmund University, Otto-Hahn-Str. 12, 44319 Dortmund,}\\ \normalsize{christian.janiesch@tu-dortmund.de}\\
\normalsize{$^{4}$FAU Erlangen-Nürnberg, Lange Gasse 20, 90403 Nürnberg,}\\ \normalsize{patrick.zschech@fau.de }\\
\normalsize{$^\ast$To whom correspondence should be addressed; e-mail:  feuerriegel@lmu.de.}
}

\date{}

\begin{document} 

\baselineskip24pt


\maketitle 

\begin{biseabstract}
The term ``generative AI'' refers to computational techniques that are capable of generating seemingly new, meaningful content such as text, images, or audio from training data. The widespread diffusion of this technology with examples such as Dall-E 2, GPT-4, and Copilot is currently revolutionizing the way we work and communicate with each other. In this article, we provide a conceptualization of generative AI as an entity in socio-technical systems and provide examples of models, systems, and applications. Based on that, we introduce limitations of current generative AI and provide an agenda for Business \& Information Systems Engineering (BISE) research. Different from previous works, we focus on generative AI in the context of information systems, and, to this end, we discuss several opportunities and challenges that are unique to the BISE community and make suggestions for impactful directions for BISE research.
\end{biseabstract}

\begin{bisekeywords}
Generative AI; Artificial intelligence; Decision support; Content creation; Information systems
\end{bisekeywords}


\section{Introduction}
\label{sec:intro}


Tom Freston is credited with saying ``Innovation is taking two things that exist and putting them together in a new way''. For a long time in history, it has been the prevailing assumption that artistic, creative tasks such as writing poems, creating software, designing fashion, and composing songs could only be performed by humans. This assumption has changed drastically with recent advances in artificial intelligence (AI) that can generate new content in ways that cannot be distinguished anymore from human craftsmanship.


The term \emph{generative AI} refers to computational techniques that are capable of generating seemingly new, meaningful content such as text, images, or audio from training data. The widespread diffusion of this technology with examples such as Dall-E 2, GPT-4, and Copilot is currently revolutionizing the way we work and communicate with each other. Generative AI systems can not only be used for artistic purposes to create new text mimicking writers or new images mimicking illustrators, but they can and will assist humans as intelligent question-answering systems. Here, applications include information technology (IT) help desks where generative AI supports transitional knowledge work tasks and mundane needs such as cooking recipes and medical advice. Industry reports suggest that generative AI could raise global gross domestic product (GDP) by $7\%$ and replace $300$ million jobs of knowledge workers \citep{goldman_sachs_2023}. Undoubtedly, this has drastic implications not only for the Business \& Information Systems Engineering (BISE) community, where we will face revolutionary opportunities, but also challenges and risks that we need to tackle and manage to steer the technology and its use in a responsible and sustainable direction.


In this Catchword article, we provide a conceptualization of generative AI as an entity in socio-technical systems and provide examples of models, systems, and applications. Based on that, we introduce limitations of current generative AI and provide an agenda for BISE research. Previous papers discuss generative AI around specific methods such as language models \citep[e.g.,][]{teubnerwelcome2023,Dwivedi.2023,schobel_charting_2023} or specific applications such as marketing \citep[e.g.,][]{peres2023chatgpt}, innovation management \citep{burger_use_2023}, scholarly research \citep[e.g.,][]{DBLP:journals/isr/isred23,DBLP:journals/isj/isjed23}, and education \citep[e.g.,][]{kasneci2023chatgpt,gimpel2023unlocking}. Different from these works, we focus on generative AI in the context of information systems, and, to this end, we discuss several opportunities and challenges that are unique to the BISE community and make suggestions for impactful directions for BISE research.

\section{Conceptualization}
\label{sec:concept}

\subsection{Mathematical Principles of Generative AI}

Generative AI is primarily based on generative modeling, which has distinctive mathematical differences from discriminative modeling \citep{ng_discriminative_2001} often used in data-driven decision support. In general, discriminative modeling tries to separate data points $X$ into different classes $Y$ by learning decision boundaries between them (e.g., in classification tasks with $Y \in \{ 0, 1 \}$). In contrast to that, generative modeling aims to infer some actual data distribution. Examples can be the joint probability distribution $P(X, Y)$ of both the inputs and the outputs or $P(Y)$, but where $Y$ is typically from some high-dimensional space. By doing so, a generative model offers the ability to produce new synthetic samples (e.g., generate new observation-target-pairs $(X,Y)$ or new observations $X$ given a target value $Y$) \citep{bishop_pattern_2006}.

\begin{figure}[H]
\centering
\includegraphics[width=.85\textwidth]{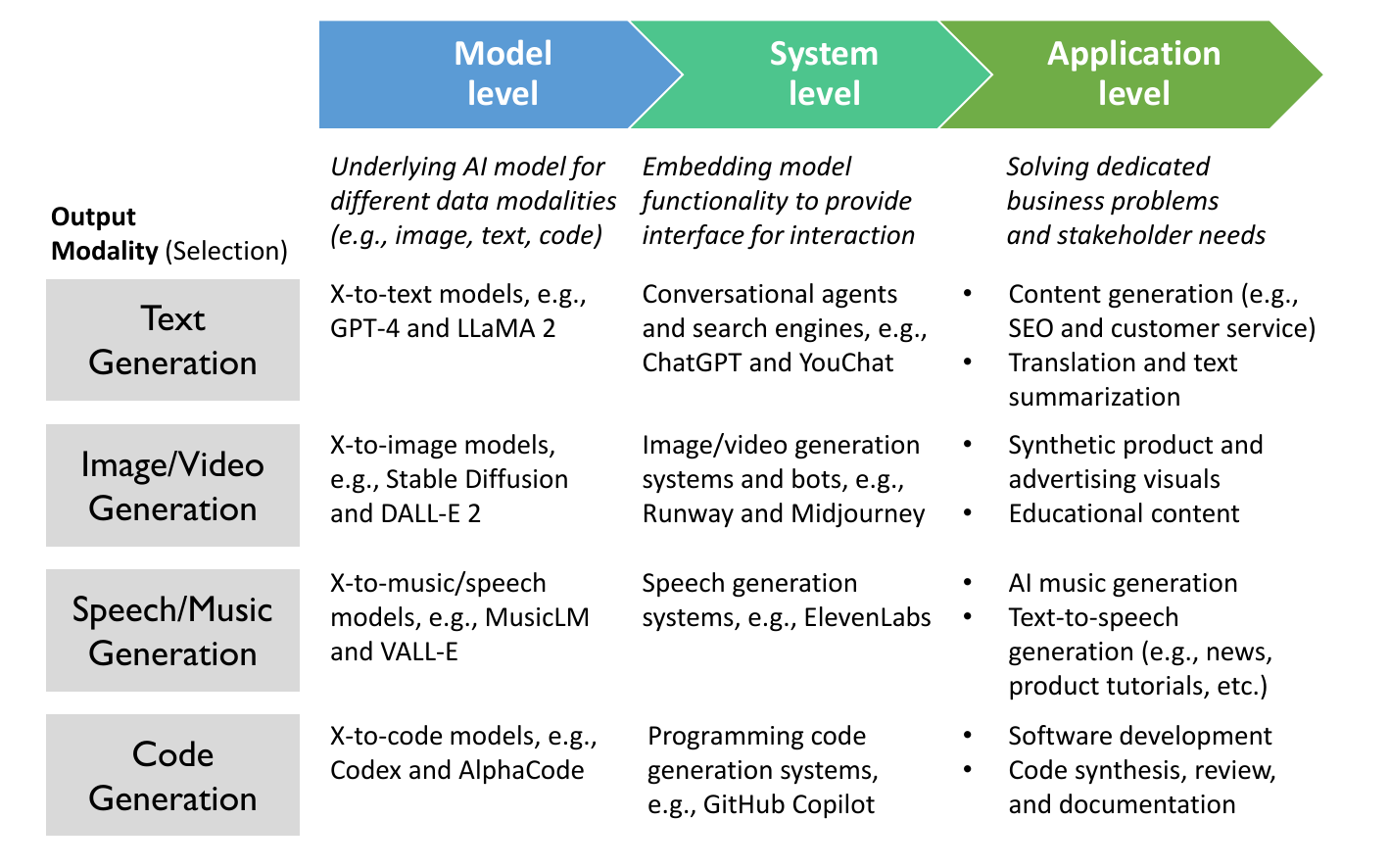}
\caption{A model-, system-, and application-level view on generative AI.}
\label{fig:Framework}
\end{figure}

Building upon the above, a \emph{generative AI model} refers to generative modeling that is instantiated with a machine learning architecture (e.g., a deep neural network) and, therefore, can create new data samples based on learned patterns.\footnote{It should be noted, however, that advanced generative AI models are often not based on a single modeling principle or learning mechanism, but combine different approaches. For example, language models from the GPT family first apply a generative pre-training stage to capture the distribution of language data using a language modeling objective, while downstream systems typically then apply a discriminative fine-tuning stage to adapt the model parameters to specific tasks (e.g., document classification, question answering). Similarly, ChatGPT combines techniques from generative modeling together with discriminatory modeling and reinforcement learning (see \Cref{fig:training}).} Further, a \emph{generative AI system} encompasses the entire infrastructure, including the model, data processing, and user interface components. The model serves as the core component of the system, which facilitates interaction and application within a broader context. Lastly, \emph{generative AI applications} refer to the practical use cases and implementations of these systems, such as search engine optimization (SEO) content generation or code generation that solve real-world problems and drive innovation across various domains. \Cref{fig:Framework} shows a systematization of generative AI across selected data modalities (e.g., text, image, and audio) and the model-, system-, and application-level perspectives, which we detail in the following section.

Note that the modalities in \Cref{fig:Framework} are neither complete nor entirely distinctive and can be detailed further. In addition, many unique use cases such as, for example, modeling functional properties of proteins \citep{unsal2022learning} can be represented in another modality such as text.

\subsection{A Model-, System-, and Application-Level View of Generative AI}
\label{sec:model_system_application}

\subsubsection{Model-Level View}
\label{sec:model}

A generative AI model is a type of machine learning architecture that uses AI algorithms to create novel data instances, drawing upon the patterns and relationships observed in the training data. A generative AI model is of critically central yet incomplete nature, as it requires further fine-tuning to specific tasks through systems and applications.

Deep neural networks are particularly well suited for the purpose of data generation, especially as deep neural networks can be designed using different architectures to model different data types \citep{janiesch_machine_2021, kraus_deep_2020}, for example, sequential data such as human language or spatial data such as images. Table~\ref{tab:glossary} presents an overview of the underlying concepts and model architectures that are common in the context of generative AI, such as diffusion probabilistic models for text-to-image generation or the transformer architecture and (large) language models (LLMs) for text generation. GPT (short for generative pre-trained transformer), for example, represents a popular family of LLMs, used for text generation, for instance, in the conversational agent ChatGPT.

Large generative AI models that can model output in and across specific domains or specific data types in a comprehensive and versatile manner are oftentimes also called \emph{foundation models} \citep{Bommasani2021FoundationModels}. Due to their size, they exhibit two key properties: \emph{emergence}, meaning the behavior is oftentimes implicitly induced rather than explicitly constructed (e.g., GPT models can create calendar entries in the .ical format even though such models were not explicitly trained to do so), and \emph{homogenization}, where a wide range of systems and applications can now be powered by a single, consolidated model (e.g., Copilot can generate source code across a wide range of programming languages).

Figure~\ref{fig:Framework} presents an overview of generative AI models along different, selected data modalities, which are pre-trained on massive amounts of data. Note that we structure the models in Figure~\ref{fig:Framework} by their output modality such as X-to-text or X-to-image. For example, GPT-4 as the most recent generative AI model underlying OpenAI's popular conversational agent ChatGPT \citep{openai2023gpt4} accepts both image and text inputs to generate text outputs. Similarly, Midjourney accepts both modalities to generate images. To this end, generative AI models can also be grouped into unimodal and multimodal models. Unimodal models take instructions from the same input type as their output (e.g., text). On the other hand, multimodal models can take input from different sources and generate output in various forms. Multimodal models exist across a variety of data modalities, for example for text, image, and audio. Prominent examples include Stable Diffusion \citep{rombach2022high} for text-to-image generation, MusicLM \citep{agostinelli2023musiclm} for text-to-music generation, Codex \citep{chen2021evaluating} and AlphaCode \citep{li2022competition} for text-to-code generation, and as mentioned above GPT-4 for image-to-text as well as text-to-text generation \citep{openai2023gpt4}.

\begin{figure}
\centering
\begin{tabular}{c}
\multicolumn{1}{l}{\hspace{2cm}\Large\textbf{\textsf{a}}} \\
\includegraphics[width=.6\textwidth]{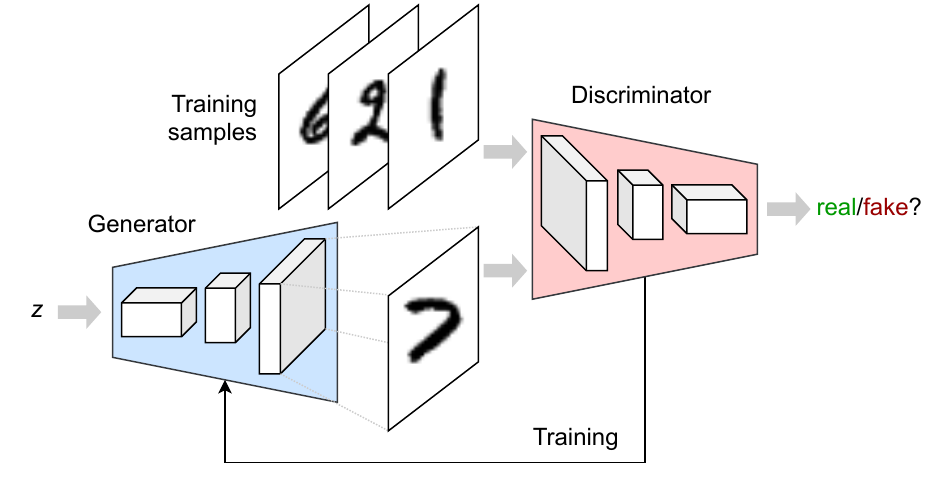} \\[1cm]
\multicolumn{1}{l}{\hspace{0cm}\Large\textbf{\textsf{b}}} \\
\includegraphics[width=.85\textwidth]{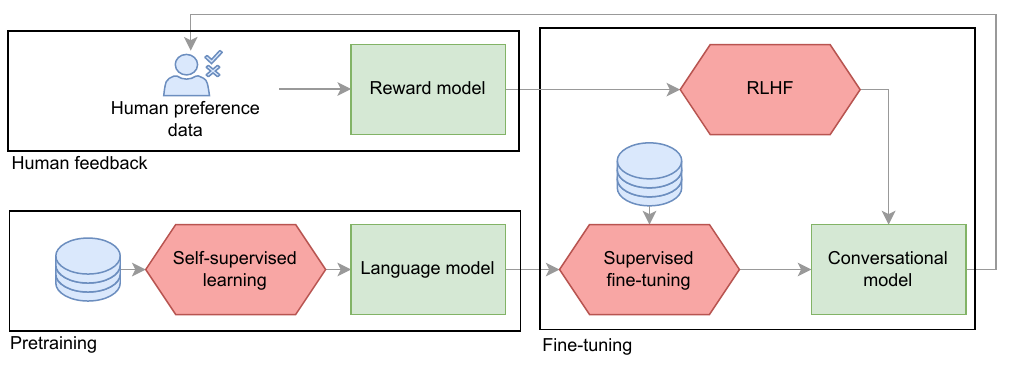} 
\end{tabular}
\caption{Examples of different training procedures for generative AI models. (a)~Generative adversarial network (GAN) where $z$ is random input. (b)~Reinforcement learning from human feedback (RLHF) as used in conversational generative AI models.}
\label{fig:training}
\end{figure}

The underlying training procedures vary greatly across different generative AI models (see Figure~\ref{fig:training}). For example, generative adversarial networks (GANs) are trained through two competing objectives \citep{GAN2014}, where one is to create new synthetic samples while the other tries to detect synthetic samples from the actual training samples, so that the distribution of synthetic samples is eventually close to the distribution of the training samples. Differently, systems such as ChatGPT-based conversational models use reinforcement learning from human feedback (RLHF). RLHF as used by ChatGPT proceeds in three steps to first create demonstration data for prompts, then to have users rank the quality of different outputs for a prompt, and finally to learn a policy that generates desirable output via reinforcement learning so that the output would score well during ranking \citep{ziegler2019fine}.

\begin{table}[H]
\centering
\tiny
\renewcommand{\arraystretch}{1.2}
\begin{tabular}{p{2.7cm} p{12.3cm}}
\toprule
\textbf{Concept} & \textbf{Description} \\
\midrule
Diffusion probabilistic models & Diffusion probability models are a class of latent variable models that are common for various tasks such as image generation \citep{ho2020denoising}. Formally, diffusion probability models capture the image data by modeling the way data points diffuse through a latent space, which is inspired by statistical physics. Specifically, they typically use Markov chains trained with variational inference and then reverse the diffusion process to generate a natural image. A notable variant is Stable Diffusion \citep{rombach2022high}. Diffusion probability models are also used in commercial systems such as DALL-E and Midjourney.\\ 

Generative adversarial network & A GAN is a class of neural network architecture with a custom, adversarial learning objective \citep{GAN2014}. A GAN consists of two neural networks that contest with each other in the form of a zero-sum game, so that samples from a specific distribution can be generated. Formally, the first network $G$ is called the generator, which generates candidate samples. The second network $D$ is called the discriminator, which evaluates how likely the candidate samples come from a desired distribution. Thanks to the adversarial learning objective, the generator learns to map from a latent space to a data distribution of interest, while the discriminator distinguishes candidates produced by the generator from the true data distribution (see Figure~\ref{fig:training}).\\

(Large) language model & A (large) language model (LLM) refers to neural networks for modeling and generating text data that typically combine three characteristics. First, the language model uses a large-scale, sequential neural network (e.g., transformer with an attention mechanism). Second, the neural network is pre-trained through self-supervision in which auxiliary tasks are designed to learn a representation of natural language without risk of overfitting (e.g., next-word prediction). Third, the pre-training makes use of large-scale datasets of text (e.g., Wikipedia, or even multi-language datasets). Eventually, the language model may be fine-tuned by practitioners with custom datasets for specific tasks (e.g., question answering, natural language generation). Recently, language models have evolved into so-called LLMs, which combine billions of parameters. Prominent examples of massive LLMs are BERT \citep{devlin2018bert} and GPT-3 \citep{brown2020language} with $\sim$340 million and $\sim$175 billion parameters, respectively.\\

Reinforcement learning from human feedback & RLHF learns sequential tasks (e.g., chat dialogues) from human feedback. Different from traditional reinforcement learning, RLHF directly trains a so-called reward model from human feedback and then uses the model as a reward function to optimize the policy, which is optimized through data-efficient and robust algorithms \citep{ziegler2019fine}. RLHF is used in conversational systems such as ChatGPT \citep{ChatGPT} for generating chat messages, such that new answers accommodate the previous chat dialogue and ensure that the answers are in alignment with predefined human preferences (e.g., length, style, appropriateness).\\

Prompt learning & Prompt learning is a method for LLMs that uses the knowledge stored in language models for downstream tasks \citep{liu2023pre}. In general, prompt learning does not require any fine-tuning of the language model, which makes it efficient and flexible. A prompt is a specific input to a language model (e.g., ``The movie was superb. Sentiment: \textvisiblespace``) and then the most probable output $s \in \{ \textrm{``positive''}, \textrm{``negative''} \}$ instead of the space is picked. Recent advances allow for more complex data-driven prompt engineering, such as tuning prompts via reinforcement learning \citep{liu2023pre}.\\ 

seq2seq & The term sequence-to-sequence (seq2seq) refers to machine learning approaches where an input sequence is mapped onto an output sequence \citep{sutskever2014sequence}. An example is machine learning-based translation between different languages. Such seq2seq approaches consist of two main components: An encoder turns each element in a sequence (e.g., each word in a text) into a corresponding hidden vector containing the element and its context. The decoder reverses the process, turning the vector into an output element (e.g., a word from the new language) while considering the previous output to model dependencies in language. The idea of seq2seq models has been extended to allow for multi-modal mappings such as text-to-image or text-to-speech mappings.\\

Transformer & A transformer is a deep learning architecture \citep{vaswani2017attention} that adopts the mechanism of self-attention which differentially weights the importance of each part of the input data. Like recurrent neural networks (RNNs), transformers are designed to process sequential input data, such as natural language, with applications for tasks such as translation and text summarization. However, unlike RNNs, transformers process the entire input all at once. The attention mechanism provides context for any position in the input sequence. Eventually, the output of a transformer (or an RNN in general) is a document embedding, which presents a lower-dimensional representation of text (or other input) sequences where similar texts are located in closer proximity which typically benefits downstream tasks as this allows to capture semantics and meaning \citep{siebers_survey_2022}.\\

Variational autoencoder & A variational autoencoder (VAE) is a type of neural network that is trained to learn a low-dimensional representation of the input data by encoding it into a compressed latent variable space and then reconstructing the original data from this compressed representation. VAEs differ from traditional autoencoders by using a probabilistic approach to the encoding and decoding process, which enables them to capture the underlying structure and variation in the data and generate new data samples from the learned latent space \citep{kingma_auto_encoding_2022}. This makes them useful for tasks such as anomaly detection and data compression but also image and text generation.\\ 

Zero-shot learning / few-shot learning & Zero-shot learning and few-shot learning refer to different paradigms of how machine learning deals with the problem of data scarcity. Zero-shot learning is when a machine is taught how to learn a task from data without ever needing to access the data itself, while few-short learning refers to when there are only a few specific examples. Zero-shot learning and few-shot learning are often desirable in practice as they reduce the cost of setting up AI systems. LLMs are few-shot or zero-shot learners \citep{brown2020language} as they just need a few samples to learn a task (e.g., predicting the sentiment of reviews), which makes LLMs highly flexible as a general-purpose tool.\\
\bottomrule
\end{tabular}
\caption{Glossary of key concepts in generative AI. }
\label{tab:glossary}
\end{table}

\subsubsection{System-Level View}
\label{sec:system}

Any system consists of a number of elements that are interconnected and interact with each other. For generative AI systems, this comprises not only the aforementioned generative AI model but also the underlying infrastructure, user-facing components, and their modality as well as the corresponding data processing (e.g., for prompts). An example would be the integration of deep learning models, like Codex \citep{chen2021evaluating}, into a more interactive and comprehensive system, like GitHub Copilot, which allows its users to code more efficiently. Similarly, Midjourney's image generation system builds on an undisclosed X-to-image generation model that users can interact with to generate images using Discord bots. Thus, generative AI systems embed the functionality of the underlying mathematical model to provide an interface for user interaction. This step augments the model-specific capabilities, enhancing its practicability and usability across real-world use cases. 

Core concerns when embedding deep learning models in generative AI systems generally are scalability (e.g., distributed computing resources), deployment (e.g., in various environments and for different devices), and usability (e.g., a user-friendly interface and intent recognition). As pre-trained open-source alternatives to closed-source, proprietary models continue to be released, making these models available to their users (be it companies or individuals) becomes increasingly important. For both open-source and closed-source models, unexpected deterioration of model performance over time highlights the need for continuous model monitoring \citep{chen2023chatgpt}. Although powerful text-generating models existed before the release of the ChatGPT system in November 2022, ChatGPT's ease of use also for non-expert users was a core contributing factor to its explosive worldwide adoption.

Moreover, on the system level, multiple components of a generative AI system can be integrated or connected to other systems, external databases with domain-specific knowledge, or platforms. For example, common limitations in many generative AI models are that they were trained on historical data with specific cut-off date and thus do not store information beyond or that an information compression takes place because of which generative AI models may not remember everything that they saw during training \citep{chiang_2023}. Both limitations can be mitigated by augmenting the model with functionality for real-time information retrieval, which can substantially enhance its accuracy and usefulness. Relatedly, in the context of text generation, online language modeling addresses the problem of outdated models by continuously training them on up-to-date data.\footnote{See https://github.com/huggingface/olm-datasets for a script that enables users to pull up-to-date data from the web for training online language models, for instance, from Common Crawl and Wikipedia.} Thereby, such models can then be knowledgeable of recent events that their static counterparts would not be aware of due to their training cut-off dates.

\subsubsection{Application-Level View}
\label{sec:application}

Generative AI applications are generative AI systems situated in organizations to deliver value by solving dedicated business problems and addressing stakeholder needs. They can be regarded as human-task-technology systems or information systems that use generative AI technology to augment human capacities to accomplish specific tasks. This level of generative AI encompasses countless real-world use cases: These range from SEO content generation \citep{reisenbichler2022frontiers}, over synthetic movie generation \citep{metz_2023} and AI music generation \citep{garcia_2023}, to natural language-based software development \citep{chen2021evaluating}. 

Generative AI applications will give rise to novel technology-enabled modes of work. The more users will familiarize themselves with these novel applications, the more they will trust or mistrust them as well as use or disuse them. Over time, applications will likely transition from mundane tasks such as writing standard letters and getting a dinner reservation to more sensitive tasks such as soliciting medical or legal advice. They will involve more consequential decisions, which may even involve moral judgment \citep{krugel2023chatgpt}. This ever-increasing scope and pervasiveness of generative AI applications give rise to an imminent need not only to provide prescriptions and principles for trustworthy and reliable designs, but also for scrutinizing the effects on the user to calibrate qualities such as trust appropriately. The (continued) use and adoption of such applications by end users and organizations entails a number of fundamental socio-technical considerations to descry innovation potential and affordances of generative AI artifacts.

\subsection{A Socio-Technical View on Generative AI}

As technology advances, the definition and extent of what constitutes AI are continuously refined, while the reference point of human intelligence stays comparatively constant \citep{DBLP:journals/misq/BerenteN21}. With generative AI, we are approaching a further \emph{point of refinement}. In the past, the capability of AI was mostly understood to be analytic, suitable for decision-making tasks. Now, AI gains the capability to perform generative tasks, suitable for content creation. While the procedure of content creation to some respect can still be considered analytic as it is inherently probabilistic, its results can be creative or even artistic as generative AI combines elements in novel ways. Further, IT artifacts were considered passive as they were used directly by humans. With the advent of agentic IT artifacts \citep{DBLP:journals/misq/BairdM21} powered by LLMs \citep{park2023generative}, this human agency primacy assumption needs to be revisited and impacts how we devise the relation between human and AI based on their potency. Eventually, this may require \emph{AI capability models} to structure, explain, guide, and constrain the different abilities of AI systems and their uses as AI applications.

Focusing on the interaction between humans and AI, so far, for analytic AI, the concept of delegation has been discussed to establish a hierarchy for decision-making \citep{DBLP:journals/misq/BairdM21}. With generative AI, a human uses prompts to engage with an AI system to create content, and the AI then interprets the humans' intentions and provides feedback to presuppose further prompts. At first glance, this seems to follow a delegation pattern as well. Yet, the subsequent process does not, as the output of the AI can be suggestive to the other and will inform their further involvement directly or subconsciously. Thus, the process of creation rather follows a co-creation pattern, that is, the practice of collaborating in different roles to align and offer diverse insights to guide a design process \citep{DBLP:journals/jbr/RamaswamyV18}. Using the lens of agentic AI artifacts, initiation is not limited to humans.

The abovementioned interactions also impact our current understanding of hybrid intelligence as the integration of humans and AI, leveraging the unique strengths of both. Hybrid intelligence argues to address the limitations of each intelligence type by combining human intuition, creativity, and empathy with the computational power, accuracy, and scalability of AI systems to achieve enhanced decision-making and problem-solving capabilities \citep{DBLP:journals/bise/DellermannESL19}. With generative AI and the AI's capability to co-create, the understanding of what constitutes this collective intelligence begins to shift. Hence,  novel \emph{human-AI interaction models and patterns} may become necessary to explain and guide the behavior of humans and AI systems to enable effective and efficient use in AI applications on the one hand and, on the other hand, to ensure envelopment of AI agency and reach \citep{DBLP:journals/jais/AsatianiMNPRS21}.

On a theoretical level, this shift in human-computer or rather human-AI interaction fuels another important observation: The theory of mind is an established theoretical lens in psychology to describe the cognitive ability of individuals to understand and predict the mental states, emotions, and intentions of others \citep{DBLP:journals/cogsci/CarlsonS13,Baron-Cohen1997-BARMAE-8, gray2007dimensions}. This skill is crucial for social interactions, as it facilitates empathy and allows for effective communication. Moreover, conferring a mind to an AI system can substantially drive usage intensity \citep{hartmann2023mind}. The development of a theory of mind in humans is unconscious and evolves throughout an individual's life. The more natural AI systems become in terms of their interface and output, the more a theory of mind for human-computer interactions becomes necessary. Research is already investigating how AI systems can become theory-of-mind-aware to better understand their human counterpart \citep{DBLP:conf/icml/RabinowitzPSZEB18,DBLP:conf/comphci/Celikok18}. However, current AI systems hardly offer any cues for interactions. Thus, humans are rather void of a theory to explain their understanding of intelligent behavior by AI systems, which becomes even more important in a co-creation environment that does not follow a task delegation pattern. A \emph{theory of the artificial mind} that explains how individuals perceive and assume the states and rationale of AI systems to better collaborate with them may alleviate some of these concerns.

\section{Limitations of Current Generative AI}
\label{sec:limitations}

In the following, we discuss four salient boundaries of generative AI that, we argue, are important limitations in real-world applications. The following limitations are of technical nature in that they refer to how current generative AI models make inferences, and, hence, the limitations arise at the model level. Because of this, it is likely that limitations will persist in the long run, with system- and application-level implications. 

\textbf{Incorrect outputs.} Generative AI models may produce output with errors. This is owed to the underlying nature of machine learning models relying on probabilistic algorithms for making inferences. For example, generative AI models generate the most probable response to a prompt, not necessarily the correct response. As such, challenges arise as, by now, outputs are indistinguishable from authentic content and may present misinformation or deceive users \citep{spitale2023ai}. In LLMs, this problem in emergent behavior is called hallucination \citep{ji2023survey}, which refers to mistakes in the generated text that are semantically or syntactically plausible but are actually nonsensical or incorrect. In other words, the generative AI model produces content that is not based on any facts or evidence, but rather on its own assumptions or biases. Moreover, the output of generative AI, especially that of LLMs, is typically not easily verifiable.

The correctness of generative AI models is highly dependent on the quality of training data and the according learning process. Generative AI systems and applications can implement correctness checks to inhibit certain outputs. Yet, due to the black-box nature of state-of-the-art AI models \citep{rai2020explainable}, the usage of such systems critically hinges on users' trust in reliable outputs. The closed source of commercial off-the-shelf generative AI systems aggravates this fact and prohibits further tuning and re-training of the models. One solution for addressing the downstream implications of incorrect outputs is to use generative AI to produce explanations or references, which can then be verified by users. However, such explanations are again probabilistic and thus subject to errors; nevertheless, they may help users in their judgment and decision-making when to accept outputs of generative AI and when not. 

\textbf{Bias and fairness.} Societal biases permeate everyday human-generated content \citep{eskreis2022surprised}. The unbiasedness of vanilla generative AI is very much dependent on the quality of training data and the alignment process. Training deep learning models on biased data can amplify human biases, replicate toxic language, or perpetuate stereotypes of gender, sexual orientation, political leaning, or religion \citep[e.g.,][]{caliskan2017semantics,hartmann2023political}. Recent studies expose the harmful biases embedded in multimodal generative AI models such as CLIP (contrastive language-image pre-training; \cite{wolfe2022evidence}) and the CLIP-filtered LAION dataset \citep{birhane2021multimodal}, which are core components of generative AI models (e.g., Dall-E 2 or Stable Diffusion). Human biases can also creep into the models in other stages of the model engineering process. For instruction-based language models, the RLHF process is an additional source of bias \citep{openai2023bias}. Careful coding guidelines and quality checks can help address these risks.

Addressing bias and thus fairness in AI receives increasing attention in the academic literature 
\citep{dolata2022sociotechnical,schramowski2022large,ferrara2023should,DeArteaga.2022,Feuerriegel.2020,vonzahn2022}, but remains an open and ongoing research question. For example, the developers of Stable Diffusion flag ``probing and understanding the limitations and biases of generative models'' as an important research area \citep{rombach2022high}. Some scholars even attest to models certain moral self-correcting capabilities \citep{ganguli2023capacity}, which may attenuate concerns of embedded biases and result in more fairness. In addition, on the system and application level, mitigation mechanisms can be implemented to address biases embedded in the deep learning models and create more diverse outputs (e.g., updating the prompts ``under the hood'' as done by Dall-E 2 to increase the demographic diversity of the outputs). Yet, more research is needed to get closer to the notion of fair AI.

\textbf{Copyright violation.} Generative AI models, systems, and applications may cause a violation of copyright laws because they can produce outputs that resemble or even copy existing works without permission or compensation to the original creators \citep{smits2022generative}. Here, two potential infringement risks are common. On the one hand, generative AI may make illegal copies of a work, thus violating the reproduction right of creators. Among others, this may happen when a generative AI was trained on original content that is protected by copyright but where the generative AI produces copies. Hence, a typical implication is that the training data for building generative AI systems must be free of copyrights. Crucially, copyright violation may nevertheless still happen even when the generative AI has never seen a copyrighted work before, such as, for example, when it simply produces a trademarked logo similar to that of Adidas but without having never seen that logo before. On the other hand, generative AI may prepare derivative works, thus violating the transformation right of creators. To this end, legal questions arise around the balance of originality and creativity in generative AI systems. Along these lines, legal questions also arise around who holds the intellectual property for works (including patents) produced by a generative AI.

\textbf{Environmental concerns.} Lastly, there are substantial environmental concerns from developing and using generative AI systems due to the fact that such systems are typically built around large-scale neural networks, and, therefore, their development and operation consume large amounts of electricity with immense negative carbon footprint \citep{schwartz2020green}. For example, the carbon emission for training a generative AI model such as GPT-3 was estimated to have produced the equivalent of 552~t~CO$_2$ and thus amounts to the annual CO$_2$ emissions of several dozens of households \citep{khan2021ai}. Owing to this, there are ongoing efforts in AI research to make the development and deployment of AI algorithms more carbon-friendly, through more efficient training algorithms, through compressing the size of neural network architectures, and through optimized hardware \citep{schwartz2020green}. 

\section{Implications and Future Directions for the BISE Community} 
\label{sec:research_agenda}

In this section, we draw a number of implications and future research directions which, on the one hand, are of direct relevance to the BISE community as an application-oriented, socio-technical research discipline and, on the other hand, offer numerous research opportunities, especially for BISE researchers due to their interdisciplinary background. We organize our considerations according to the individual departments of the BISE journal (see Table~\ref{tbl:future_directions} for an overview of exemplary research questions).

\begin{table}
\centering
\footnotesize
\begin{tabular}{p{5cm}p{9cm}}
\toprule
BISE department & Research questions (examples) \\
\midrule
Business process management & \vspace{-0.45cm}\begin{itemize}
    \item How can generative AI assist in automating routine tasks?
    \item How can generative AI reveal process innovation opportunities and support process (re-)design initiatives? 
\end{itemize} \\
Decision analytics and data science  & \vspace{-0.45cm}\begin{itemize}
    \item How can generative AI models be effectively fine-tuned for domain-specific applications? 
    \item How can the reliability of generative AI systems be improved? 
\end{itemize} \\
Digital business management and digital leadership  & \vspace{-0.45cm}\begin{itemize}
    \item How can generative AI support managerial tasks such as resource allocation? 
    \item How will the digital work of employees change with smart assistants powered by generative AI? 
\end{itemize} \\
Economics of information systems  & \vspace{-0.45cm}\begin{itemize}
    \item What are the welfare implications of generative AI? 
    \item Which jobs and tasks are affected most by generative AI?
\end{itemize} \\
Enterprise modeling and enterprise engineering & \vspace{-0.45cm}\begin{itemize}
    \item How can generative AI be used to support the construction and maintenance of enterprise models?
    \item How can generative AI support in enterprise applications (e.g., CRM, BI, etc.)?
\end{itemize} \\
Human computer interaction and social computing & \vspace{-0.45cm}\begin{itemize}
    \item How should generative AI systems be designed to foster trust? 
    \item What countermeasures are effective to prevent users from falling for AI-generated disinformation? 
    \item To what extent can generative AI replace or augment crowdsourcing tasks? 
    \item How can generative AI assist in education? 
\end{itemize} \\
Information systems engineering and technology & \vspace{-0.45cm}\begin{itemize}
    \item What are effective design principles for developing generative AI systems?
    \item How can generative AI support design science projects to foster creativity in the development of new IT artifacts?
\end{itemize} \\[-0.4cm]
\bottomrule
\end{tabular}
\caption{Examples of research questions for future BISE research on generative AI.}
\label{tbl:future_directions}
\end{table}

\subsection{Business Process Management}
\label{sec:bpm}

Generative AI will have a strong impact on the field of Business Process Management (BPM) as it can assist in automating routine tasks, improving customer and employee satisfaction, and revealing process innovation opportunities \citep{beverungen_seven_2021}, especially in creative processes \citep{haase_hanel_2023}. Concrete implications and research directions can be connected to various phases of the BPM lifecycle model \citep{vidgof_large_2023}. For example, in the context of process discovery, generative AI models could be used to generate process descriptions, which can help businesses identify and understand the different stages of a process \citep{kecht_quantifying_2023}. From the perspective of business process improvement, generative process models could be used for idea generation and to support innovative process (re-)design initiatives \citep{van_dun_processgan_2023}. In this regard, there is great potential for generative AI to contribute to both exploitative as well as explorative BPM design strategies \citep{grisold_five_2022}. In addition, natural language processing tasks related to BPM such as process extraction from text could benefit from generative AI without further fine-tuning using prompt engineering \citep{busch_2023}. Likewise, other phases can benefit owing to generative AI's ability to learn complex and non-linear relationships in dynamic business processes that can be used for implementation as well as in simulation and predictive process monitoring among other things.

In the short term, robotic process automation \citep{van_der_aalst_robotic_2018,herm2021symbolic} will benefit as formerly handcrafted processing rules can not only be replaced, but entirely new types of automation can be enabled by retrofitting and thus intelligentizing legacy software. In the long run, we also see a large potential to support the phase of business process execution in traditional BPM. Specifically, we anticipate the development of a new generation of process guidance systems. While traditional system designs are based on static and manually-crafted knowledge bases \citep{morana_designing_2019}, more dynamic and adaptive systems are feasible on the basis of large enterprise-wide trained language models. Such systems could improve knowledge retrieval tasks from a wide variety of heterogeneous sources, including manuals, handbooks, e-mails, wikis, job descriptions, etc. This opens up new avenues of research into how unstructured and distributed organizational knowledge can be incorporated into intelligent process guidance systems. 

\subsection{Decision Analytics and Data Science}

Despite the huge progress in recent years, several analytical and technical questions around the development of generative AI have yet to be solved. One open question relates to how generative AI can be effectively customized for domain-specific applications and thus improve performance through higher degrees of contextualization. For example, novel and scalable techniques are needed to customize conversational agents based on generative AI for applications in medicine or finance. This will be crucial in practice to solve specific BISE-related tasks where customization may bring additional performance gains. Novel techniques for customization must be designed in a way that ensures the safety of proprietary data and prevents the data from being disclosed. Moreover, new frameworks are needed for prompt engineering that are designed from a user-centered lens and thus promote interpretability and usability. 

Another important research direction is to improve the reliability of generative AI systems. For example, algorithmic solutions are needed on how generative AI can detect and mitigate hallucination. In addition to algorithmic solutions, more effort is also needed to develop user-centered solutions, that is, how users can reduce the risk of falling for incorrect outcomes, for example, by developing better ways how outputs can be verified (e.g., by offering additional explanations or references).

Finally, questions arise about how generative AI can natively support decision analytics and data science projects by closing the gap between modeling experts and domain users \citep{zschech_intelligent_2020}. For instance, it is commonly known that many AI models used in business analytics are difficult to understand by non-experts \citep[cf.][]{senoner2022using}. As a remedy, generative AI could be used to generate descriptions that explain the logic of business analytics models and thus make the decision logic more intelligible. One promising direction could be, for example, to use generative AI for translating post~hoc explanations derived from approaches like SHAP or LIME into more intuitive textual descriptions or generate user-friendly descriptions of models that are intrinsically interpretable \citep{slack_explaining_2023, zilker_best_2023}.

\subsection{Digital Business Management and Digital Leadership}

Generative AI has great potential to contribute to different types of value creation mechanisms, including knowledge creation, task augmentation, and autonomous agency. However, this also requires the necessary organizational capabilities and conditions, where further research is needed to examine these ingredients more closely for the context of generative AI to steer the technological possibilities in a successful direction \citep{shollo_shifting_2022}.

That is, generative AI will lead to the development of new business ideas, unseen product and service innovations, and ultimately to the emergence of completely new business models. At the same time, it will also have a strong impact on intra-organizational aspects, such as work patterns, organizational structures, leadership models, and management practices. In this regard, we see that AI-based assistant systems previously centered around desktop automation taking over more and more routine tasks such as event management, resource allocation, and social media account management to free up even more human capacity \citep{maedche_ai-based_2019}. Further, in algorithmic management \citep{benlian_algorithmic_2022, cameron_algorithmic_2023}, it should be examined how existing theories and frameworks need to be contextualized or fundamentally extended in light of the increasingly powerful capabilities of generative AI. 

However, there are not only implications at the management level. The future of work is very likely to change at all levels of an organization \citep{feuerriegel2022bringing}. Due to the multi-modality of generative AI models, it is conceivable that employees will work increasingly via smart, speech-based interfaces, whereby the formulation of prompts and the evaluation of their results could become a key activity. Against this background, it is worth investigating which new competencies are required to handle this emerging technology \citep[cf.][]{debortoli_comparing_2014} and which entirely new job profiles, such as prompt engineers, may evolve in the near future \citep{strobelt_interactive_2023}.

Generative AI is also expected to fundamentally reform the way organizations manage, maintain, and share knowledge. Referring to the sketched vision of a new process guidance system in \Cref{sec:bpm}, we anticipate a number of new opportunities for digital knowledge management, among others automated knowledge discovery based on large amounts of unstructured distributed data (e.g., identification of new product combinations), improved knowledge sharing by automating the process of creating, summarizing, and disseminating content (e.g., automated creation of wikis and FAQs in different languages), and personalized knowledge delivery to individual employees based on their specific needs and preferences (e.g., recommendations for specific training material).

\subsection{Economics of Information Systems}

Generative AI will have significant economic implications across various industries and markets. Generative AI can increase efficiency and productivity by automating many tasks that were previously performed by humans, such as content creation, customer service, code generation, etc. This can reduce costs and open up new opportunities for growth and innovation \citep{eloundou_gpts_2023}. For example, AI-based translation between different languages is responsible for significant economic gains \citep{brynjolfsson2019does}. The BISE community can contribute by providing quantification through rigorous causal evidence. Given the velocity of AI research, it may be necessary to take a more abstract problem view instead of a concrete tool view.  For example, BISE research could run field experiments to compare programmers with and without AI support and thereby assess whether generative AI systems for coding can improve the speed and quality of code development. Similarly, researchers could test whether generative AI will make artists more creative as they can more easily create new content. A similar pattern was previously observed for AlphaGo, which has led humans to become better players in the board game Go \citep{shin2023superhuman}.

Generative AI is likely to transform the industry as a whole. This may hold true in the case of platforms that make user-generated content available (e.g., shutterstock.com, pixabay.com, stackoverflow.com), which may be replaced by generative AI systems. Here, further research questions arise as to whether the use of generative AI can lead to a competitive advantage and how generative AI changes competition. For example, what are the economic implications if generative AI is developed as open-source vs. in closed-source systems? In this regard, a salient success factor for the development of conversational agents based on generative AI (e.g., ChatGPT) are data from user interactions through dialogues and feedback on whether the dialog was helpful. Hence, the value of such interaction data is poorly understood and what it means if such data are only available to a few Big Tech companies. 

The digital transformation from generative AI also poses challenges and opportunities for economic policy. It may affect future work patterns and, indirectly, worker capability via restructured learning mechanisms. It may also affect content sharing and distribution and, hence, have non-trivial implications on the exploitation and protection of intellectual properties. On top of that, a growing concentration of power over AI innovation in the hands of a few companies may result in a monopoly of AI capabilities and hamper future innovation, fair competition, scientific progress, and thus welfare and human development at large. All of these future impacts are important to understand and provide meaningful directions for shaping economic policy.

\subsection{Enterprise Modeling and Enterprise Engineering}

Enterprise models are important artifacts for capturing insights into the core components and structures of an organization, including business processes, resources, information flows, and IT systems \citep{vernadat_enterprise_2020}. A major drawback of traditional enterprise models is that they are static and may not provide the level of abstraction that is required by the end user. Likewise, their construction and maintenance are time-consuming and expensive and require manual effort and human expertise \citep{silva_maintenance_2021}. With generative AI, we see a large potential that many of these limitations can be addressed by generative AI as assistive technology \citep{DBLP:journals/bise/SandkuhlFHKMOSU18}, for example by automatically creating and updating enterprise models at different levels of abstraction or generating multi-modal representations. 

First empirical results suggest that generative AI is able to generate useful conceptual models based on textual problem descriptions. \cite{fill_conceptual_2023} show that ER, BPMN, UML, and Heraklit models can not only be generated with very high to perfect accuracy from textual descriptions, but they also explored the interpretation of existing models and received good results. In the near future, we expect more research that deals with the development, evaluation, and application of more advanced approaches. Specifically, we expect that learned representations of enterprise models can be transformed into more application-specific formats and can either be enriched with further details or reduced to the essential content.

Against this background, the concept of ``digital twins'', virtual representations of enterprise assets, may experience new accentuation and extensions \citep{dietz_digital_2020}. Especially, in the public sector, where most organizational assets are non-tangible in the form of defined services, specified procedures, legal texts, manuals, and organizational charts, generative AI can play a crucial role in digitally mirroring and managing such assets along their lifecycles. Similar benefits could be explored with physical assets in Industry 4.0 environments \citep{lasi_industry_2014}.

In enterprise engineering, the role of generative AI systems in existing as well as newly emerging IT landscapes to support the business goals and strategies of an organization gives rise to numerous opportunities (e.g., in office solutions, customer relationship management and business analytics applications, knowledge management systems, etc.). Generative AI systems have the potential to evolve into core enterprise applications that can either be hosted on-premise or rented in the cloud. Unsanctioned use bears the risk that third-party applications will be used for job-related tasks without explicit approval or even knowledge of the organization. This phenomenon is commonly known as shadow IT and theories and frameworks have been proposed to explain this phenomenon, as well as recommending actions and policies to mitigate associated risks \citep[cf.][]{haag_shadow_2017, klotz_von_2022}. In the light of generative AI, however, such approaches have to be revisited for their applicability and effectiveness and, if necessary, need to be extended. Nevertheless, this situation also offers the potential to explore and design new approaches for more effective API management (e.g., including novel app store solutions, privacy and security mechanisms, service level definitions, pricing, and licensing models) so that generative AI solutions can be smoothly integrated into existing enterprise IT infrastructures without risking any unauthorized use and confidentiality breaches. 

\subsection{Human Computer Interaction and Social Computing}

Salient behavioral questions related to the interactions between humans and generative AI systems are still unanswered. Examples are related to the perception, acceptance, adoption, and trust of systems using generative AI. A study found that news was believed less if generated by generative AI instead of humans \citep{longoni2022news} and another found that there is a replicant effect \citep{DBLP:conf/chi/JakeschFMHN19}. Such behavior is likely to be context-specific and will vary by other antecedents highlighting the need for a principled theoretical foundation to build successful generative AI systems. The BISE community is well positioned to develop rigorous design recommendations.

Further, generative AI is a key enabler for developing high-quality interfaces for information systems based on natural language that promote usability and accessibility. For example, such interfaces will not only make interactions more intuitive but will also facilitate people with disabilities. Generative AI is likely to increase the ``degree of intelligence'' of user assistance systems. However, the design of effective interactions must also be considered when increasing the degree of intelligence \citep{maedche2016advanced}. Similarly, generative AI will undoubtedly have an impact on (computer-mediated) communication and collaboration, such as within companies. For example, generative AI can create optimized content for social media, emails, and reports. It can also help to improve the onboarding of new employees by creating personalized and interactive training materials. It can also enhance collaboration within teams by providing creative and intelligence conservation agents that suggest, summarize, and synthesize information based on the context of the team (e.g., automated meeting notes).   

Several applications and research opportunities are related to the use of generative AI in marketing and, especially, e-commerce. It is expected that generative AI can automate the creation of personalized marketing content, for instance, different sales slogans for introverts vs. extroverts \citep{matz2017psychological} or other personality traits as personalized marketing content is more effective than a one-content-fits-all approach \citep{matz2023potential}. Generative AI may automate various tasks in marketing and media where content generation is needed (e.g., writing news stories, summarizing web pages for mobile devices, creating thumbnail images for news stories, translating written news to audio for blind people and Braille-supported formats for deaf people) that may be studied in future research. Moreover, generative AI may be used in recommender systems to boost the effectiveness of information dissemination through personalization as content can be tailored better to the abilities of the recipient. 

The education sector is another example that will need to reinvent in some parts following the availability of conversational agents \citep{kasneci2023chatgpt,gimpel2023unlocking}. At first glance, generative AI seems to constitute an unauthorized aid that jeopardizes student grading so far relying on written examinations and term papers. However, over time, examinations will adapt, and generative AI will enable the development of comprehensive digital teaching assistants as well as the creation of supplemental teaching material such as teaching cases and recap questions. Further, the educator's community will need to develop novel guidelines and governance frameworks that educate learners to rely appropriately on generative AI systems, how to verify model outputs, and to engineer prompts rather than the output itself.

In addition, generative AI, specifically, LLMs, can not only be used to spot harmful content on social media \citep[e.g.,][]{maarouf2023hqp}, but it can also create realistic disinformation (e.g., fake news, propaganda) that is hard to detect by humans \citep{kreps2022all, jakesch2023human}. Notwithstanding, AI-generated disinformation has previously evolved as so-called deepfakes \citep{mirsky2021creation}, but recent advances in generative AI reduce the cost of creating such disinformation and allow for unprecedented personalization. For example, generative AI can automatically adapt the tone and narrative of misinformation to specific audiences that identify as extroverts or introverts, left- or right-wing partisans, or people with particular religious beliefs. 

Lastly, generative AI can facilitate -- or even replace -- traditional crowdsourcing where annotations or other knowledge tasks are handled by a larger pool of crowd workers, for example in social media content annotation \citep{gilardi_chatgpt_2023} or market research on willingness-to-pay for services and products \citep{brand2023using}. In general, we expect that generative AI will automate many other tasks being a zero-shot / few-shot learner. However, this may also unfold negative implications: Users may contribute less to question-answering forums such as stackoverflow.com, which thus may reduce human-based knowledge creation impairing the future performance of AI-based question-answering systems that rely upon human question-answering content for training. In a similar vein, the widespread availability of generative AI systems may also propel research around virtual assistants. Previously, research made use of ``Wizard-of-Oz'' experiments \citep{diederich2020designing}, while future research may build upon generative AI systems instead. 

Crucially, automated content generation using generative AI is a new phenomenon, but automation in general and how people are affected by automated systems has been studied by scholars for decades. Thus, existing theories on the interplay of humans with automated systems may be contextualized to generative AI systems.
 
\subsection{Information Systems Engineering and Technology}

Generative AI offers many engineering- and technology-oriented research opportunities for the Information Systems community as a design-oriented discipline. This includes developing and evaluating design principles for generative AI systems and applications to extend the limiting boundaries of this technology (cf. \Cref{sec:limitations}). As such, design principles can focus on how generative AI systems can be made explainable to enable interpretability, understanding, and trust; how they can be designed reliable to avoid discrimination effects or privacy issues; and how they can be built more energy efficient to promote environmental sustainability \citep[cf.][]{schoormann_artificial_2023}. While a lot of research is already being conducted in technology-oriented disciplines such as computer science, the BISE community can add its strength by looking at design aspects through a socio-technical lens, involving individuals, teams, organizations, and societal groups in design activities, and thereby driving the field forward with new insights from a human-machine perspective \citep{maedche_ai-based_2019}. 

Further, we see great potential that generative AI can be leveraged to improve current practices in design science research projects when constructing novel IT artifacts \citep[see][]{hevner_roles_2019}. Here, one of the biggest potentials could lie in the support of knowledge retrieval tasks. Currently, design knowledge in the form of design requirements, design principles, and design features is often only available in encapsulated written papers or implicitly embedded in instantiated artifacts. Generative AI has the potential to extract such design knowledge that is spread over a broad body of interdisciplinary research and make it available in a collective form for scholars and practitioners. This could also overcome the limitation that design knowledge is currently rarely reused, which hampers the fundamental idea of knowledge accumulation in design science research \citep{schoormann_how_2021}.

Besides engineering actual systems and applications, the BISE community should also investigate how generative AI can be used to support creativity-based tasks when initiating new design projects. In this regard, a promising direction could be to incorporate generative AI in design thinking and similar methodologies to combine human creativity with computational creativity \citep{hawlitschek_interview_2023}. This may support different phases and steps of innovation projects, such as idea generation, user needs elicitation, prototyping, design evaluation, and design automation, in which different types of generative AI models and systems could be used and combined with each other to form applications for creative industries (e.g., generated user stories with textual descriptions, visual mock-ups for user interfaces, and quick software prototypes for proofs-of-concept). If generative AI is used to co-create innovative outcomes, it may also enable better reflection of the different design activities to ensure the necessary learning \citep{schoormann_act_2023}.

\section{Conclusion}
\label{sec:conclusion}

Generative AI is a branch of AI that can create new content such as texts, images, or audio that increasingly often cannot be distinguished anymore from human craftsmanship. For this reason, generative AI has the potential to transform domains and industries that rely on creativity, innovation, and knowledge processing. In particular, it enables new applications that were previously impossible or impractical for automation, such as realistic virtual assistants, personalized education and service, and digital art. As such, generative AI has substantial implications for BISE practitioners and scholars as an interdisciplinary research community. In our Catchword article, we offered a conceptualization of the principles of generative AI along a model-, system-, and application-level view as well as a social-technical view and described limitations of current generative AI. Ultimately, we provided an impactful research agenda for the BISE community and thereby highlight the manifold affordances that generative AI offers through the lens of the BISE discipline.

\section*{Acknowledgments}

During the preparation of this Catchword, we contacted all current department editors at BISE to actively seek their feedback on our suggested directions. We gratefully acknowledge their support.


\setstretch{1.0}
\bibliographystyle{apalike}
\bibliography{literature}

\end{document}